\documentclass{article}

\usepackage[final, nonatbib]{neurips_2018}

\usepackage[utf8]{inputenc} 
\usepackage[T1]{fontenc}    
\usepackage{hyperref}       
\usepackage{url}            
\usepackage{booktabs}       
\usepackage{amsfonts}       
\usepackage{nicefrac}       
\usepackage{microtype}      

\usepackage{amsmath}
\usepackage{color}
\usepackage{amssymb}
\usepackage{bbm}
\usepackage{graphicx}
\graphicspath{{fig/}}
\usepackage{subcaption}
\usepackage{arydshln}
\newdimen\figrasterwd
\figrasterwd\textwidth
\usepackage{hyperref} 
\usepackage{wrapfig}

\newcommand\model{\text{CP-AAN}}

\def\ddefloop#1{\ifx\ddefloop#1\else\ddef{#1}\expandafter\ddefloop\fi}

\def\ddef#1{\expandafter\def\csname bb#1\endcsname{\ensuremath{\mathbb{#1}}}}
\ddefloop ABCDEFGHIJKLMNOPQRSTUVWXYZ\ddefloop

\def\ddef#1{\expandafter\def\csname c#1\endcsname{\ensuremath{\mathcal{#1}}}}
\ddefloop ABCDEFGHIJKLMNOPQRSTUVWXYZ\ddefloop

\def\ddef#1{\expandafter\def\csname v#1\endcsname{\ensuremath{\mathbf{#1}}}}
\ddefloop ABCDEFGHIJKLMNOPQRSTUVWXYZabcdefghijklmnopqrstuvwxyz\ddefloop

\newcommand{\qed}{\hfill\ensuremath{\square}}

\def\ddef#1{\expandafter\def\csname v#1\endcsname{\ensuremath{\mathbf{\csname #1\endcsname}}}}
\ddefloop {alpha}{beta}{gamma}{delta}{epsilon}{varepsilon}{zeta}{eta}{theta}{vartheta}{iota}{kappa}{lambda}{mu}{nu}{xi}{pi}{varpi}{rho}{varrho}{sigma}{varsigma}{tau}{upsilon}{phi}{varphi}{chi}{psi}{omega}{Gamma}{Delta}{Theta}{Lambda}{Xi}{Pi}{Sigma}{varSigma}{Upsilon}{Phi}{Psi}{Omega}{ell}\ddefloop

\newcommand{\ind}[1]{\mathbbm{1}_{[#1]}}

\newcommand{\first}[1]{\textcolor{red}{\textbf{#1}}}
\newcommand{\second}[1]{\textcolor{blue}{#1}}
\DeclareMathOperator*{\tr}{tr}

\newcommand{\etal}{\textit{et al}.}

\newcommand{\argmin}{\ensuremath{\arg\min}}
\DeclareMathOperator*{\argminn}{arg\,min}

\DeclareMathOperator*{\diag}{diag}

\title{Low-shot Learning via Covariance-Preserving Adversarial Augmentation Networks}

\author{
    Hang Gao$^1$,
    Zheng Shou$^1$,
    Alireza Zareian$^1$,
    Hanwang Zhang$^2$,
    Shih-Fu Chang$^1$ \\
    $^1$Columbia University,
    $^2$Nanyang Technological University \\
    \small{\texttt{\{hg2469, zs2262, az2407, sc250\}@columbia.edu}} \\
    \small{\texttt{hanwangzhang@ntu.edu.sg}}
}

\begin{document}

\maketitle

\begin{abstract}
    Deep neural networks suffer from over-fitting and catastrophic forgetting when trained with small data. One natural remedy for this problem is data augmentation, which has been recently shown to be effective. However, previous works either assume that intra-class variances can always be generalized to new classes, or employ naive generation
    methods to hallucinate finite examples without modeling their latent
    distributions. In this work, we propose \textit{Covariance-Preserving
    Adversarial Augmentation Networks} to overcome existing
    limits of low-shot learning. Specifically, a novel Generative Adversarial Network is designed to
    model the latent distribution of each novel class given its related base
    counterparts. Since direct estimation of novel classes can be inductively
    biased, we explicitly preserve covariance information as the ``variability''
    of base examples during the generation process. Empirical results show that
    our model can generate realistic yet diverse examples, leading to
    substantial improvements on the ImageNet benchmark over the
    state of the art.
\end{abstract}

\section{Introduction}

The hallmark of learning new concepts from very few examples characterizes human
intelligence. Though constantly pushing limits forward in various visual tasks,
current deep learning approaches struggle in cases when abundant training data
is impractical to gather. A straightforward idea to learn new concepts is to
fine-tune a model pre-trained on \textit{base} categories, using limited data from another
set of \textit{novel} categories. However, this usually leads to catastrophic forgetting
\cite{goodfellow2013empirical}, i.e., fine-tuning makes the model over-fitting on
novel classes, and agnostic to the majority of base classes
\cite{kirkpatrick2017overcoming, shmelkov2017incremental}, deteriorating overall
performance.

One way to address this problem is to augment data for novel
classes. Since generating images could be both
unnecessary \cite{xian2017feature} and impractical \cite{salimans2016improved}
on large datasets, feature augmentation \cite{arandjelovic2012three,
chen2018person} is more preferable in this scenario. Building upon learned
representations
\cite{vinyals2016matching,snell2017prototypical,sung2017learning}, recently two
variants of generative models show the promising capability of learning variation
modes from base classes to imagine the missing pattern of novel classes. Hariharan~\etal~proposed \textit{Feature Hallucination (FH)}~\cite{hariharan2016low},
which can learn a finite set of transformation mappings between examples in each base
category and directly apply them to seed novel points for extra data. However,
since mappings are \textit{enumerable} (even in large amount), this model suffers
from poor generalization. To address this issue, Wang~\etal{}~\cite{wang2018low}
proposed \textit{Feature Imagination (FI)}, a meta-learning based generation
framework that can train an agent to synthesize extra data given a specific
task. They circumvented the demand for latent distribution of novel classes by
end-to-end optimization. But the generation results usually collapse into
certain modes. Finally, it should be noted that both works erroneously assume
that intra-class variances of base classes are shareable with any novel classes.
For example, the visual variability of the concept \textit{lemon} cannot be generalized to other irrelevant
categories such as \textit{raccoon}.

\begin{wrapfigure}{r}{.5\textwidth}
    \begin{center}
    \includegraphics[width=.35\textwidth]{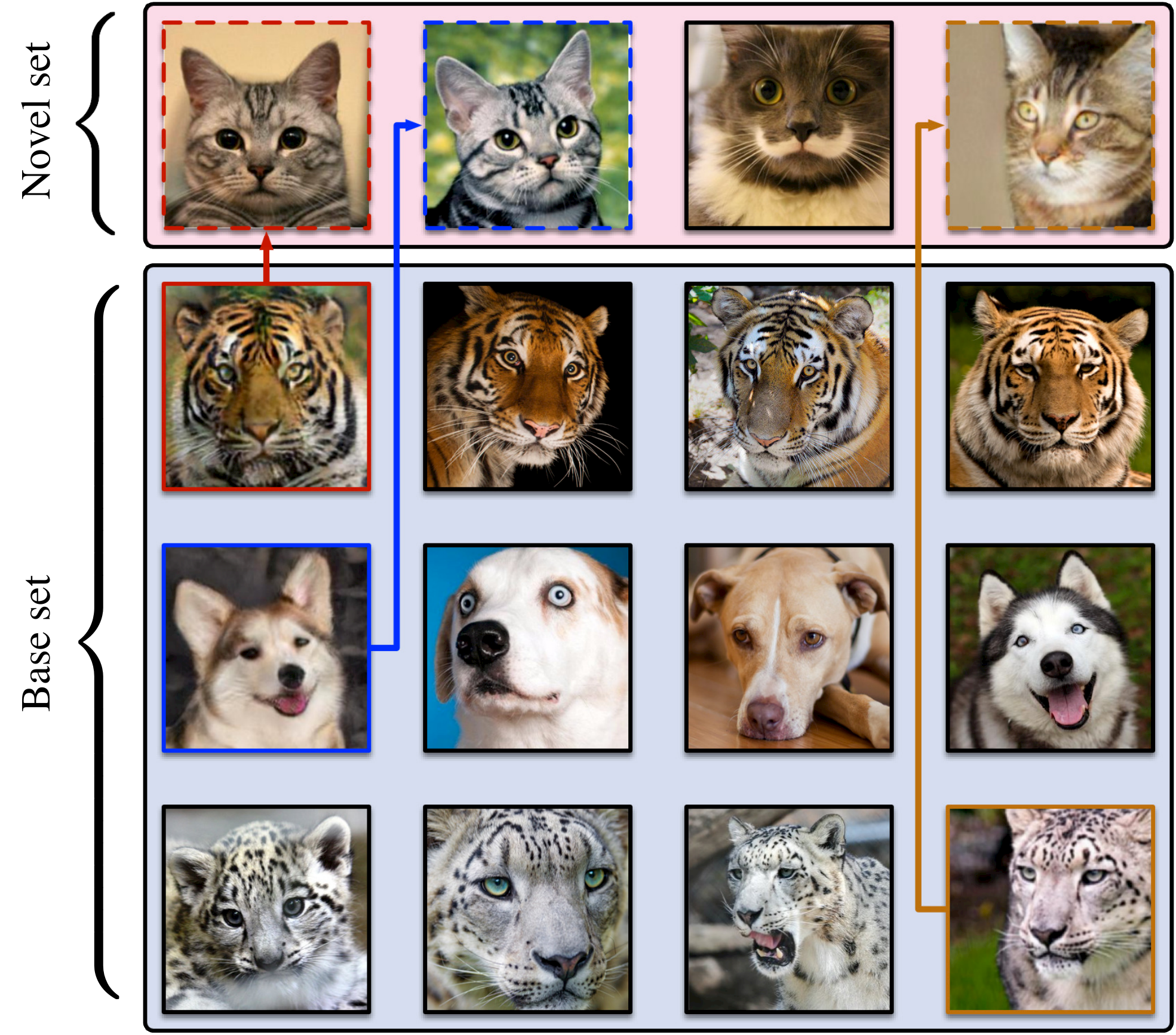}
    \end{center}
    \caption{\textbf{Conceptual illustration of our method.} Given an example
    from a novel class, we translate examples from related base classes into the target class for augmentation. 
    Image by \cite{huang2018munit}.}
\end{wrapfigure}

In this work, we propose a new approach to addressing the problem of low-shot learning by enabling
better feature augmentation beyond current limits. Our approaches are novel in two aspects: modeling 
and training strategy. We propose
\textit{Covariance-Preserving Adversarial Augmentation Networks (\model{})}, a
new class of Generative Adversarial Networks (GAN) \cite{goodfellow2014generative,
mirza2014conditional} for feature augmentation. We take inspiration from
unpaired image-to-image translation \cite{zhu2017unpaired, zhu2017toward} and
formulate our feature augmentation problem as an imbalanced set-to-set
translation problem where the conditional distribution of examples of each
novel class can be conceptually expressed as a mixture of related base classes.
We first extract all related base-novel class pairs by an intuitive yet effective approach called
\textit{Neighborhood Batch Sampling}. Then, our model aims to learn the latent
distribution of each novel class given its base counterparts. Since the direct
estimation of novel classes can be inductively biased during this process, we
explicitly preserve the covariance base examples during the generation process.

We systematically evaluate our approach by considering a series of objective functions.
Our model achieves the state-of-the-art performance over the challenging
ImageNet benchmark \cite{deng2009imagenet}. With ablation studies,
we also demonstrate the effectiveness of each component in our method.

\section{Related Works} 

\textbf{Low-shot Learning} \quad
For quick adaptation when very few novel examples are available, the community has
often used a meta-agent \cite{lei1996dynamic} to further tune base
classifiers \cite{vinyals2016matching,snell2017prototypical,sung2017learning}.
Intuitive yet often ignored, feature augmentation was recently brought
into the field by Hariharan~\etal~\cite{hariharan2016low} to ease the data scarce
scenario. Compared to traditional meta-learning based approaches, they have
reported noticeable improvement on not only the \textit{conventional} setting (i.e., to
test on novel examples only), but also the more challenging \textit{generalized} setting (i.e., to
test on all classes). Yet the drawback is that both the original work and its
variants \cite{wang2018low} fail to synthesize diverse examples because of
ill-constrained generation processes. Our
approach falls in this line of research while seeking more principal guidance
from base examples in a selective, class-specific manner.

\textbf{Generative Adversarial Network for Set-to-set Translation} \quad
GANs \cite{goodfellow2014generative} map each latent
code from an easily sampled prior to a realistic sample of a complex target distribution.
Zhu~\etal~\cite{zhu2017unpaired} have achieved astounding results on
image-to-image translation without any paired training samples. In our
case, diverse feature augmentation is feasible through conditional translation 
given a pair of related novel and base classes.
Yet two main challenges remain: practically,
not all examples are semantically translatable. Second, given extremely scarce data 
for novel classes, we are unable to estimate their latent distributions
(see Figure \ref{fig:cov-demo}). In this work, we thoroughly investigate
conditional GAN variants inspired by previous works \cite{salimans2016improved,
mirza2014conditional, zhu2017toward, odena2016conditional} to enable low-shot generation. Furthermore,
we introduce a novel batch sampling technique for learning salient set-to-set
mappings using unpaired data with categorical conditions.

\textbf{Generation from Limited Observations} \quad
Estimation of latent distribution from a handful of observations is biased and
inaccurate \cite{tobin1958estimation,candes2007dantzig}. 
The Bayesian approaches aim to model latent distributions of a variety of classes as  hierarchical Gaussian mixture \cite{salakhutdinov2012one},
or alternatively model generation as a sequential decision making process
\cite{rezende2016one}. For GANs, Gaussian mixture noise has also been
incorporated for latent code sampling \cite{gurumurthy2017deligan}. Recent works
 \cite{mroueh2017mcgan, mroueh2017fisher}
on integral probability metrics provide theoretical guidance towards the high
order feature matching. In this paper, building upon the assumption that related
classes should have similar intra-class variance, we introduce a new loss term
for preserving covariance during the translation process.

\section{Imbalanced Set-to-set Translation} \label{sec:isst}

In this section, we formulate our low-shot feature augmentation problem
under an imbalanced set-to-set translation framework. Concretely, we are given two labeled  
datasets represented
in the same $D$-dimensional semantic space: (1) a base set $\cB = \{(\vx_b,
y_b)\,|\,\vx_b \in \bbR^D, y_b \in \cY_b\}$ consisting of abundant samples and
(2) a novel set $\cN = \{(\vx_n, y_n)\,|\,\vx_n \in \bbR^D, y_n \in
\cY_n\}$ with only a handful of observations. Their discrete label spaces are
assumed to be non-overlapping, i.e., $\cY_b \cap \cY_n = \varnothing$. Our goal is to learn a mapping function $G_n: \cB \mapsto \cN$ in order to translate
examples of the base classes into novel categories. After the generation process, a
final classifier is trained using both original examples of the base classes and all (mostly
synthesized) examples of the novel classes.

\begin{figure}[tp]
    \centering
    \parbox{\figrasterwd}{
        \begin{tabular}{c:cc}
            \quad\quad\quad
            \parbox{.24\figrasterwd}{%
                \centering
                \subcaptionbox{Problem statement}
                {\includegraphics[width=\hsize]{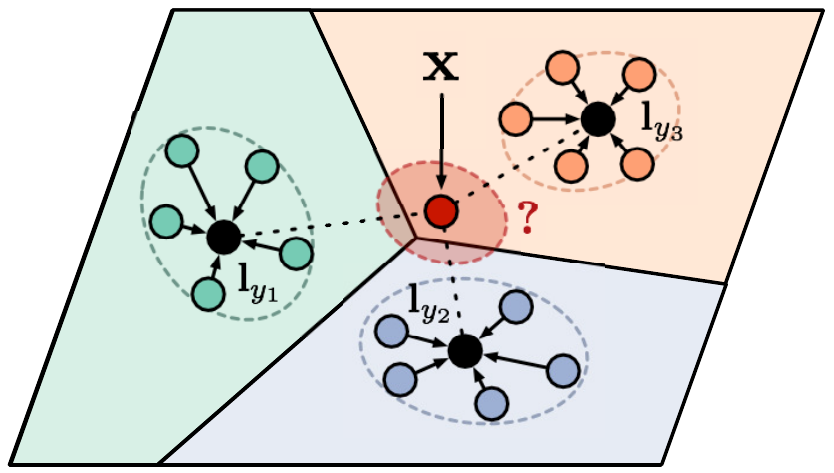}}
                \label{fig:ists}
            }
            & \parbox{.24\figrasterwd}{%
                \centering
                \subcaptionbox{Hariharan~\etal 
                \cite{hariharan2016low}}
                {\includegraphics[width=\hsize]{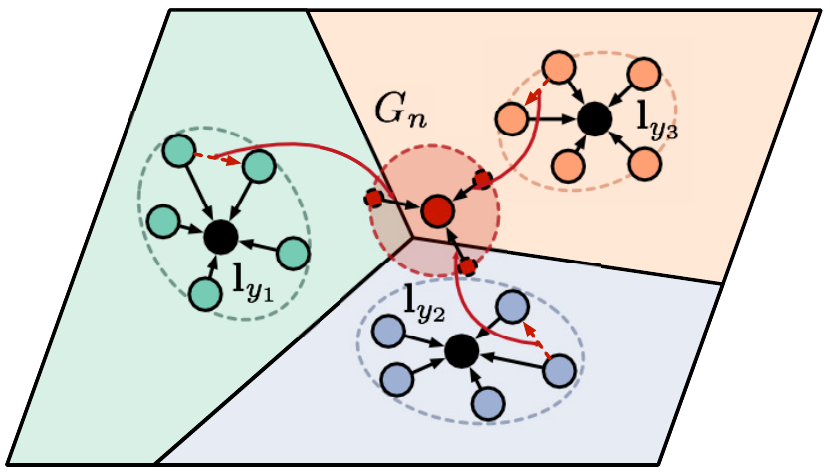}}
                \label{fig:previous}
            }
            & \parbox{.24\figrasterwd}{%
                \centering
                \subcaptionbox{Our intuitions}
                {\includegraphics[width=\hsize]{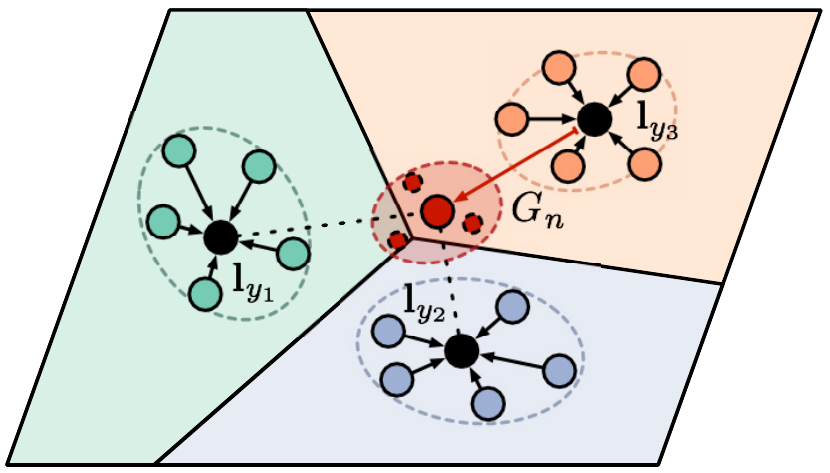}}
                \label{fig:concept}
            }
        \end{tabular}
    }
    \caption{\textbf{Imbalanced set-to-set translation and our motivations.} Examples of three base classes are visualized in a semantic space learned by Prototypical Networks
    \cite{snell2017prototypical}, along with their centroids as class prototypes. \textbf{(a):} Given a novel example $\mathbf{x}$, our goal is to
    translate base examples into the novel class, to reconstruct an estimation of the novel class distribution. \textbf{(b):} Feature Hallucination \cite{hariharan2016low} randomly
    applies transformation mappings, between sampled base pairs in the same
    class, to the seed novel example for extra data; \textbf{(c):} instead, we
    only refer to semantically similar base-novel class pairs and model the
    distribution of data for novel classes by preserving base intra-class variances.}
    \label{fig:ists-demo}
\end{figure}

Existing works \cite{hariharan2016low, wang2018low} suffer from the use of arbitrary, and thus
possibly unrelated, base classes for feature augmentation. Moreover, their performances are degraded by naive
generation methods without modeling the latent distribution of each novel class.
Our insight, conversely, is to sample extra features from continuous latent
\textit{distributions} rather than certain modes from \textit{enumerations}, by
learning a GAN model (see Figure \ref{fig:ists-demo}).

Specifically, we address two challenges that impede good
translation under imbalanced scenarios: (1) through which base-novel class pairs we
can translate; and more fundamentally, (2) through what objectives for GAN training
we can estimate the latent distribution of novel classes with limited observations. We here start by
proposing a straightforward batch sampling technique to address the first problem.
Then we suggest a simple extension of existing methods and study its
weakness, which motivates the development of our final approach.
For clarity, we introduce a toy dataset for imbalanced set-to-set translation in
Figure \ref{fig:sprl} as a conceptual demonstration of the proposed method compared to baselines.

\subsection{Neighborhood Batch Sampling} \label{sec:nbs}
It is widely acknowledged \cite{goldberger2005neighbourhood,
vinyals2016matching, snell2017prototypical} that a metric-learned high
dimensional space encodes relational semantics between examples.
Therefore, to define which base classes are translatable to a novel class,
we can rank them by their distance in a semantic space. For 
simplicity, we formulate our approach on top of Prototypical Networks
\cite{snell2017prototypical}, learned by the nearest neighbor classifier 
on the semantic space measured by the Euclidean distance. 
We represent each class $y$ as a cluster and encode
its categorical information by the cluster prototype $\vl_y \in \bbR^D$:

\begin{equation}
    \label{eq:proto}
    \vl_y = \frac{\sum_i \vx_i \cdot \ind{y_i = y}}
    {\sum_i \ind{y_i = y}}
\end{equation}

It should be noted that by ``prototype'' we mean the centroid of examples of a class. It should not be confused with the centroid of randomly sampled examples that is computed in each episode to train original Prototypical Networks.

We introduce translation mapping $R: \cY_n \mapsto \cP(\cY_b)$ where $\cP(\cY_b)$ is the
powerset of the collection of all base classes. This defines a many-to-many
relationship between novel and base classes, and is used to translate data from selected base
classes to each novel class. To this end, given a novel class $y_n$, we compute its
similarity scores $\alpha$ with all base classes $y_b$ using softmax over
Euclidean distances between prototypes, 

\begin{equation}
    \alpha(y_b, y_n) = 
    \frac{\exp{(-\| \vl_{y_b} - \vl_{y_n} \|_2^2)}}
    {\sum_{y_b' \in \cY_b} \exp{(-\| \vl_{y_b'} - \vl_{y_n} \|_2^2})}
\end{equation}

This results in a soft mapping (NBS-S) between base and novel classes, in which each novel class is paired with all base classes with soft scores. In practice, translating from all base classes is unnecessary, and computationally expensive. 
Alternatively, we consider a hard version of $R$ based on $k$-nearest neighbor search, 
where the top $k$ base classes are selected and treated as equal ($\alpha(y_b, y_n) = 1 / k$).
This hard mapping (NBS-H) saves memory, but introduces an extra hyper-parameter. 

\subsection{Adversarial Objective}
\begin{figure}[tp]
    \centering
    \begin{subfigure}[b]{0.2\textwidth}
        \includegraphics[width=\textwidth]{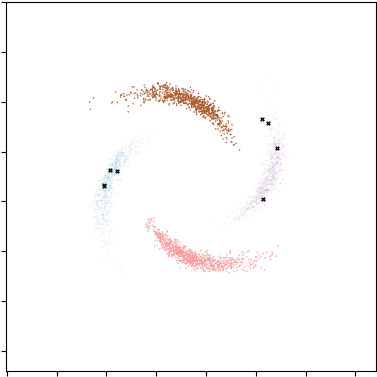}
        \caption{}
        \label{fig:sprl-raw}
    \end{subfigure}%
    \begin{subfigure}[b]{0.2\textwidth}
        \includegraphics[width=\textwidth]{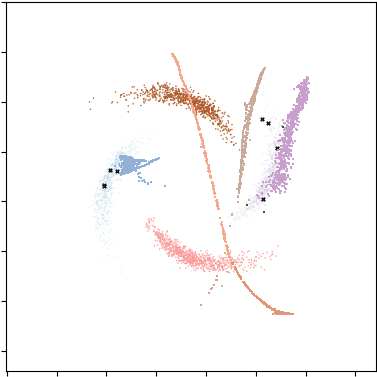}
        \caption{}
        \label{fig:sprl-cgan}
    \end{subfigure}%
    \begin{subfigure}[b]{0.2\textwidth}
        \includegraphics[width=\textwidth]{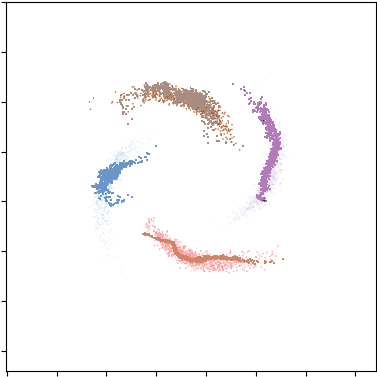}
        \caption{}
        \label{fig:sprl-ccyc-gan}
    \end{subfigure}%
    \begin{subfigure}[b]{0.2\textwidth}
        \includegraphics[width=\textwidth]{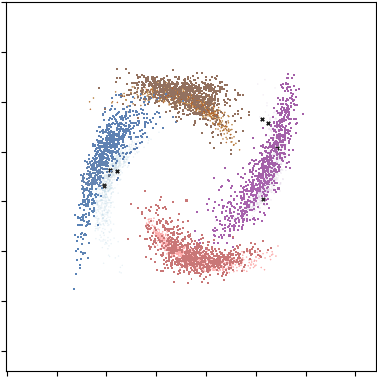}
        \caption{}
        \label{fig:sprl-deligan}
    \end{subfigure}%
    \begin{subfigure}[b]{0.2\textwidth}
        \includegraphics[width=\textwidth]{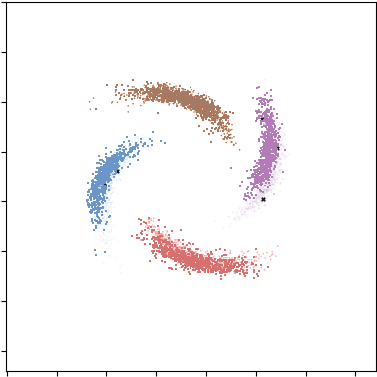}
        \caption{}
        \label{fig:sprl-ccov-gan}
    \end{subfigure}%
    \caption{\textbf{Generation results on our toy dataset.}
    \textbf{(a):} Raw distribution of the ``spiral'' dataset, which consists of two base classes (top, bottom) and two novel
    classes (left, right). Novel classes are colored with lower saturation to indicate they are not available for training. Instead, only $4$ examples (black crosses) are
    available. Generated samples are colored with higher saturation than the real data.  \textbf{(b):} \texttt{c-GAN}; note that we also show the results of translating synthesized novel samples back to original base classes with another decoupled \texttt{c-GAN} for the visual consistency with the other variants;
    \textbf{(c):}
    \texttt{cCyc-GAN}; \textbf{(d):} \texttt{cDeLi-GAN}; \textbf{(e):}
    \texttt{cCov-GAN}. Results are best viewed in color with zoom. }
    \label{fig:sprl}
\end{figure}

After constraining our translation process to selected class pairs, we develop a
baseline based on Conditional GAN (\texttt{c-GAN})
\cite{mirza2014conditional}. To this end, a discriminator $D_n$ is trained to classify real examples as the corresponding $N = |\cY_n|$ novel classes, and classify synthesized examples as an auxiliary ``fake'' class. \cite{salimans2016improved}.
The generator $G_n$ takes an example from base classes $R(y_n)$ that are paired with $y_n$ via NBS, and aims to fool the discriminator into classifying the generated example as $y_n$ instead of the ``fake''. More specifically, the adversarial objective can be written as:

\begin{align}
\label{eq:adv}
    \cL_{\text{adv}}&(G_n, D_n, \cB, \cN)
    =
    \bbE_{y_n \sim \cY_n} \bigg[
        \bbE_{\vx_n \sim \cN_{y_n}}
        \big[
            \log D_n (y_n | \vx_n)
        \big]\\
        & + \bbE_{\vx_b, y_b \sim \cB_{R(y_n)}}
        \big[
            \alpha(y_b, y_n) \log D_n (N+1 | G_n(y_n; \vx_b, y_b))
        \big]
    \bigg]
\end{align}

where $\cN_{y_n}$ consists of all novel examples labeled with $y_n$ in $\cN$ while $\cB_{R(y_n)}$ consists all base examples labeled by one of the classes in $R(y_n)$.


We train \texttt{c-GAN} by solving the minimax game of the adversarial loss.
In this scenario, there is no explicit way to incorporate base classes
intra-class variance into the generation of new novel examples. Also, any
mappings that collapse synthesized features into existing observations yield
the optimal solution \cite{goodfellow2014generative}. These facts lead to
unfavorable generation results as shown in Figure \ref{fig:sprl-cgan}. We next
explore different ways to \textit{explicitly} force the generator to learn the 
latent conditional distributions.

\subsection{Cycle-consistency Objective}
A natural idea for preventing modes from getting dropped is to apply
the cycle-consistency constraint whose effectiveness has been proven over
image-to-image translation tasks \cite{zhu2017unpaired}. Besides extra 
supervision, it eliminates the demand for paired data, which is impossible to
acquire for the low-shot learning setting. We extend this method for our
conditional scenario and derive $\texttt{cCyc-GAN}$. Specifically, we learn
two generators: $G_n$, which is our main target, and $G_b: \cN \mapsto \cB$ as an
auxiliary mapping that reinforces $G_n$.
We train the generators such that, the translation cycle recovers the original
embedding in either a forward cycle $\cN \mapsto \cB \mapsto \cN$ or a backward
cycle $\cB \mapsto \cN \mapsto \cB$. Our cycle-consistency objective could then
be derived as,

\begin{align}
    \cL_{\text{cyc}}(G_n, G_b)
    =
    &\bbE_{y_n \sim \cY_n} \bigg[
        \bbE_{\vx_n \sim \cN_{y_n},
        \vx_b, y_b \sim \cB_{R(y_n)},
        \vz \sim \cZ} \,
        \alpha(y_b, y_n) \Big[ \\
            &\| G_n(y_n; G_b(y_b; \vx_n, y_n, \vz), y_b) \|_2^2
            +
            \| G_b(y_b; G_n(y_n; \vx_b, y_b), y_n, \vz) \|_2^2
        \Big]
    \bigg]
\end{align}

where a $Z$-dimensional noise vector sampled from a distribution $\cZ$ is injected into $G_b$'s input since novel examples $x_n$ lack variability given the very limited amount of data. $\cZ$ is a normal distribution $N(0, 1)$ for our \texttt{cCyc-GAN} model.

While $G_n$ is hard to train due to the extremely small data volume; $G_b$ has more to learn from, 
and can thus indirectly guide $G_n$ through its gradient. During our
experiments, we found that cycle-consistency is indispensable for stabilizing the
training procedure. Swaminathan~\etal~\cite{gurumurthy2017deligan} observe that incorporating extra
noise from a mixture of Gaussian distributions could result in more diverse results.
Hence, we also report a variant called \texttt{cDeLi-GAN} which uses the same
objective as \texttt{cCyc-GAN}, but sample the noise vector $\vz$ from a mixture of $C$ different Gaussian distributions,

\begin{equation}
    \cZ \overset{d}{=} \frac1C \sum_{i = 1}^{C} f(\vz|\vmu_i, \vSigma_i),
    \quad \text{where} \ 
    f(\vz|\vmu, \Sigma) = 
    \frac{\exp(-\frac12 (\vz - \vmu)^T \vSigma^{-1} (\vz - \vmu))}
    {\sqrt{(2 \pi)^Z |\vSigma|}}
\end{equation}

We follow the initialization setup in the previous work \cite{gurumurthy2017deligan}.
For each $\vmu$, we sample from a uniform distribution $U(-1, 1)$. And for each $\vSigma$,
we first sample a vector $\mathbf{\sigma}$ from a Gaussian distribution $N(0, 0.2)$, then we simply set $\vSigma = \diag(\mathbf{\sigma})$

Generation results of the two aforementioned methods are shown in 
Figure \ref{fig:sprl-ccyc-gan} and
\ref{fig:sprl-deligan}. Both methods improve the diversity of generation compared
to the naive \texttt{c-GAN}, yet they either under- or over-estimate the intra-class
variance.


\subsection{Covariance-preserving Objective}
\begin{figure}[tp]
    \centering
    \begin{subfigure}[b]{0.35\textwidth}
        \includegraphics[width=\textwidth]{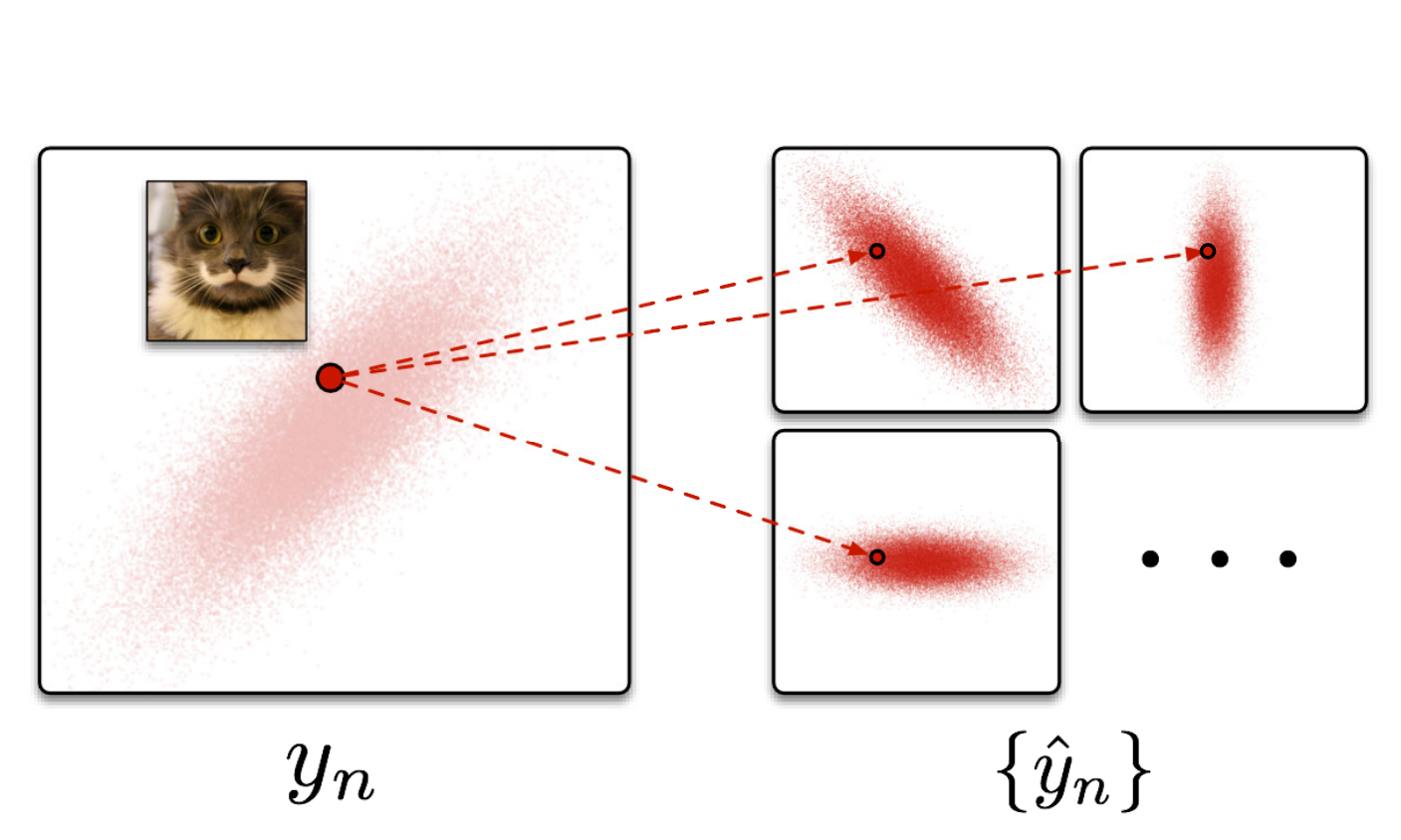}
        \caption{}
        \label{fig:cov-bad}
    \end{subfigure}
    \quad\quad
    \begin{subfigure}[b]{0.35\textwidth}
        \includegraphics[width=\textwidth]{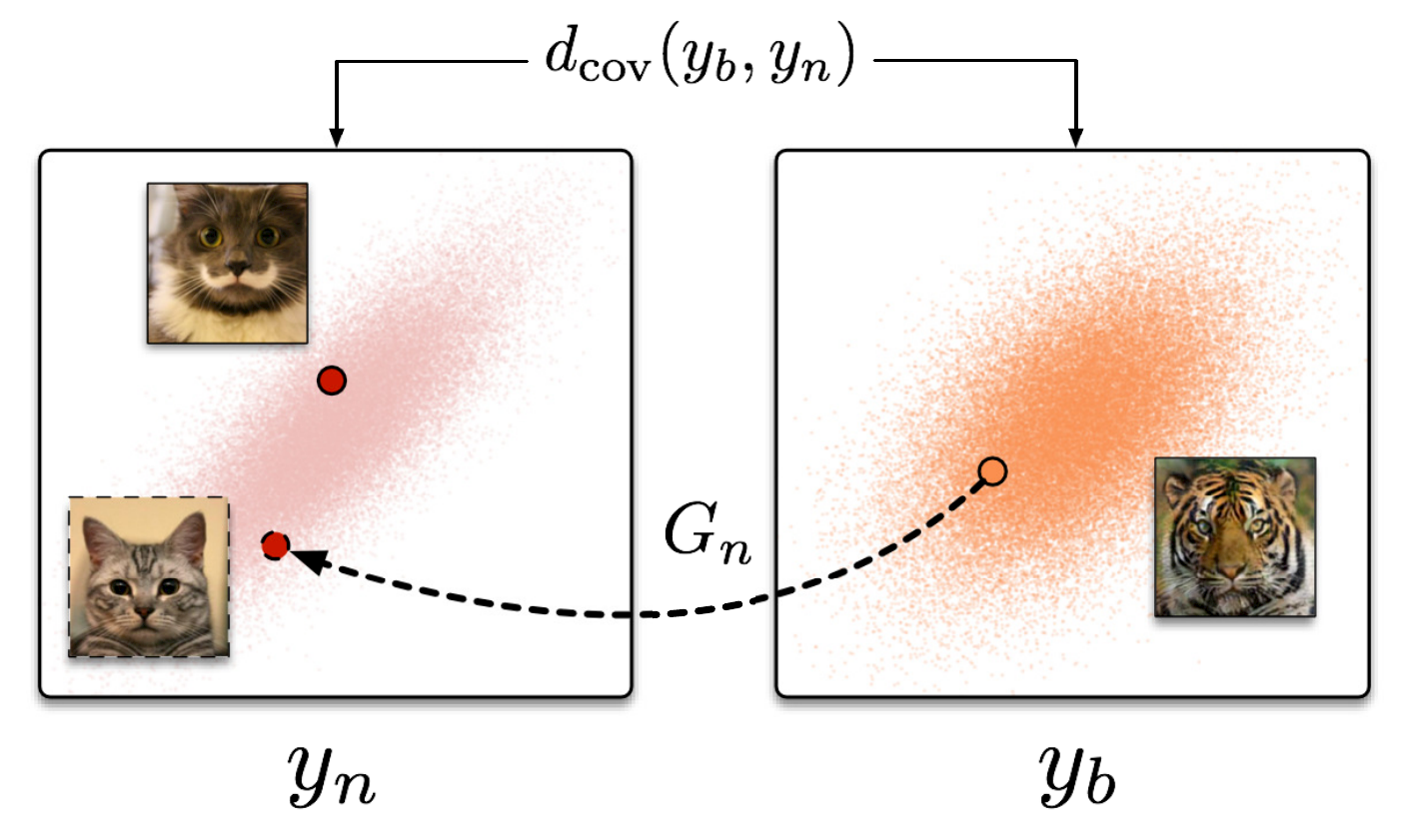}
        \caption{}
        \label{fig:cov-good}
    \end{subfigure}
    \caption{\textbf{The importance of covariance in low-shot settings.}
    Suppose we have access to a single cat image during training. \textbf{(a):}
    conventional models can easily fail since there are infinite candidate distributions that
    cannot be discriminated; \textbf{(b):} related classes should have similar
    intra-class variances. Thus, we preserve covariance information during
    translation, to transfer knowledge from base classes to novel ones.}
    \label{fig:cov-demo}
\end{figure}
While cycle-consistency alone can transfer certain degrees of intra-class
variance from base classes, we find it rather weak and unreliable since there
are still infinite candidate distributions that cannot be discriminated based on limited
observations (See Figure \ref{fig:cov-demo}).

Building upon the assumption that similar classes share similar intra-class
variance, one straightforward idea is to penalize the change of
``variability'' during translation. Hierarchical Bayesian
models \cite{salakhutdinov2012one}, prescribe each class as a multivariate Gaussian, where intra-class variability is embedded in a covariance matrix. We generalize this idea and try to maintain covariance in the translation process, although we model the class distribution by GAN instead of any prescribed distributions.

To compute the difference between two covariance matrices \cite{mroueh2017mcgan},
one typical way is to measure the worst case distance between them
using \textit{Ky Fan $m$-norm}, \textit{i.e.}, the sum of singular values of $m$-truncated
SVD, which we denote as $\| [ \cdot ]_m \|_*$. To this end, we define the pseudo-prototype $\hat{\vl}_{y_n}$ of each novel class $y_n$ as the centroid of all synthetic samples $\hat{\vx}_n = G_n(y_n; \vx_b, y_b)$ translated from related base classes. The covariance distance $d_{\text{cov}}(y_b, y_n)$ between a base-novel class pair
can then be formulated as,

\begin{equation}
    d_{\text{cov}}(y_b, y_n) = 
    \big\|
        [
            \Sigma_{\vx}(\bbP_{y_b}) - \Sigma_{G}(\bbP_{y_n})
        ]_m
    \big\|_*,
    \;\; \text{where} \ 
    \begin{cases}
        \Sigma_{\vx}(\bbP_{y}) = 
            \frac{\sum_i (\vx_i - \vl_{y_i}) (\vx_i - \vl_{y_i})^T \ind{y_i = y}}
            {\sum_i \ind{y_i = y}}
        \\
        \Sigma_{G}(\bbP_{y}) =
            \frac{\sum_j (\hat{\vx}_j - \hat{\vl}_{y_j})
                (\hat{\vx}_j - \hat{\vl}_{y_j})^T \ind{y_j = y}}
            {\sum_j \ind{y_j = y}}
    \end{cases}
\end{equation}

Consequently, our covariance-preserving objective can be written as
the expectation of the weighted covariance distance using NBS-S,

\begin{equation}
    \cL_{\text{cov}}(G_n) = 
    \bbE_{y_n \sim \cY_n} \bigg[
        \bbE_{y_b \sim R(y_n)} \big[
            \alpha(y_b, y_n) d_{\text{cov}} (y_b, y_n)
        \big]
    \bigg]
\end{equation}

Note that, for a matrix $\vX$, $\|[\vX]_m\|_*$ is non-differentiable with respect to itself, thus in practice, we calculate its subgradient instead. Specifically, we first
compute the unitary matrices $\vU$, $\vV$ by $m$-truncated SVD
\cite{xu1998truncated}, and then back-propagate $\vU \vV^T$ for sequential
parameter updates. Proof of the correctness is provided in the supplementary material.

Finally, we propose our covariance-preserving conditional cycle-GAN,
\texttt{cCov-GAN}, as:

\begin{equation}
    \label{eq:ccov-gan}
    \begin{aligned}
        G_n^* = \argmin_{G_n, G_b} \max_{D_n, D_b}
        &\cL_{\text{adv}}(G_n, D_n, \cB, \cN)
        \\
        &+ \cL_{\text{adv}}(G_b, D_b, \cN, \cB) 
        + \lambda_{\text{cyc}} \cL_{\text{cyc}}(G_n, G_b)
        + \lambda_{\text{cov}} \cL_{\text{cov}}(G_n)
    \end{aligned}
\end{equation}

As illustrated in Figure \ref{fig:sprl-ccov-gan}, preserving covariance
information from relevant base classes to a novel class can improve low-shot
generation quality. We attribute
this empirical result to the interplay of adversarial learning, cycle consistency, and covariance preservation, that respectively lead to realistic generation, semantic consistency, and diversity.

\subsection{Training}

Following recent works on meta-learning \cite{santoro2016one,hariharan2016low,wang2018low}, we design a two-stage training procedure. During the ``meta-training'' phase, we train our generative model with base examples only, by mimicking the low-shot scenario it would encounter later. After that, in the ``meta-testing'' phase, we are given novel classes as well as their low-shot examples. We use the trained $G_n$ to augment each class until it has the average capacity of the base classes. Then we train a classifier as one would normally do in a supervised setting using both real and synthesized data. For the choice of this final classifier, we apply the same one as in the original representation learning stage. For examples, we use the nearest neighbor classifier for embeddings from Prototypical Networks, and a normal linear classifier for those from ResNets.

We follow the episodic procedure used by \cite{wang2018low} during meta-training. 
In each episode, we sample $N_b$ ``meta-novel'' classes from $\cB$, 
and use the rest of $\cB$ as ``meta-base'' classes.
Then we sample $K_b$ examples from each meta-novel class as meta-novel examples.
We compute the prototypes of each class and similarity scores between each
``meta-novel'' and ``meta-base'' class. To sample a batch of size $B$, we first include 
all ``meta-novel'' examples, and sample $B - N_b \cdot K_b$ examples uniformly from the ``meta-base''
classes retrieved by translation mapping $R$.
Next, we push our samples through generations and discriminators to compute the loss.
Finally, we update their weights for the current episode and start the next one.

\section{Experiments}
This section is organized as follows. In Section \ref{sec:imagenet}, we conduct
low-shot learning experiments on the challenging
ImageNet benchmark. In Section \ref{sec:disc}, we further discuss with ablation,
both quantitatively and qualitatively, to better understand the performance
gain. We demonstrate our model's capacity to generate diverse and reliable
examples and its effectiveness in low-shot classification.

\textbf{Dataset} \quad
We evaluate our method on the real-world benchmark proposed by Hariharan~\etal~\cite{hariharan2016low}. This is a challenging task because it requires us to learn a large variety of ImageNet
\cite{deng2009imagenet} given a few exemplars for each novel classes. To this end, our model must be able to model the visual diversity of a wide range of categories and transfer knowledge between them without confusing unrelated classes. Following \cite{hariharan2016low}, we split the 1000 ImageNet classes into four disjoint
class sets $\mathcal{Y}_b^{test}, \mathcal{Y}_n^{test}, \mathcal{Y}_b^{val},
\mathcal{Y}_n^{val}$, which consist of 193, 300, 196, 311 classes respectively.
All of our parameter tuning is done on validation splits, while final results
are reported using held-out test splits.

\textbf{Evaluation} \quad
We repeat sampling novel examples five times for held-out novel sets and report
results of mean top-5 accuracy in both \textit{conventional} low-shot learning
(LSL, to test on novel classes only) and its \textit{generalized} setting (GLSL,
to test on all categories including base classes).

\textbf{Baselines} \quad 
We compare our results to the exact numbers reported
by Feature Hallucination \cite{hariharan2016low} and Feature Imagination
\cite{wang2018low}. We also compared to other
non-generative methods including classical Siamese Networks
\cite{koch2015siamese}, Prototypical Networks \cite{snell2017prototypical},
Matching Networks \cite{vinyals2016matching}, and MAML \cite{finn2017model} as well as
more recent Prototypical Matching Networks \cite{wang2018low} and Attentive Weight
Generators \cite{gidaris2018dynamic}. 
For stricter comparison, we 
provide two extra baselines to exclude the bias induced by different embedding
methods: P-FH builds on Feature Hallucinating by substituting their
non-episodic representation with learned prototypical features. Another
baseline (first row in Table \ref{tab:imagenet}), on the contrary, replaces prototypical features with raw ResNet-10 embeddings.
The results for MAML and SN are reported using their published codebases online. 

\textbf{Implementation details} \quad 
Our implementation is based on PyTorch \cite{paszke2017automatic}. Since deeper
networks would unsurprisingly result in better performance, we confine all
experiments in a ResNet-10 backbone\footnote{Released on
\url{https://github.com/facebookresearch/low-shot-shrink-hallucinate}} with a 512-d output layer. We fine-tune the backbone following the procedure described in \cite{hariharan2016low}. For all generators, we use three-layer MLPs with
all hidden layers' dimensions fixed at 512 as well as their output for
synthesized features. Our discriminators are accordingly designed as three-layer
MLPs to predict probabilities over target classes plus an extra
fake category. We use leaky ReLU of slope 0.1 without batch
normalization. Our GAN models are trained for 100000 episodes
by ADAM \cite{kingma2014adam} with initial learning rate fixed at 0.0001 which
anneals by 0.5 every 20000 episodes. We fix the hyper-parameter $m = 10$ for computing
truncated SVD. For loss term contributions, we
set $\lambda_{\text{cyc}} = 5$ and $\lambda_{\text{cov}} = 0.5$ for all final
objectives. We choose $Z = 100$ as the dimension of noise vectors for $G_b$'s input, and $C = 50$ for the Gaussian mixture. We inject prototype embeddings instead of one-hot vectors as
categorical information for all networks (prototypes for novel classes are computed using the low-shot examples only). We empirically set batch size $B = 1000$, and $N_b = 20$ and $K_b = 10$ for all training, no matter what would the number of shots be in the test. This is more efficient but possibly less accurate than \cite{snell2017prototypical} who
trained separate models for each testing scenario, so the number of shots in train and test always match. All hyper-parameters are cross-validated 
on the validation set using a coarse grid search.

\subsection{Main Results} \label{sec:imagenet}

\begin{table}[t]
    \caption{Low-shot classification top-5 accuracy${\%}$ of all comparing
    methods under LSL and GLSL settings on ImageNet dataset. All results are
    averaged over five trials separately, and omit standard deviation for all
    numbers are of the order of $0.1{\%}$. The \first{best} and \second{second
    best} methods under each setting are marked in according formats.}
    \label{tab:imagenet}
    \centering
    \resizebox{\textwidth}{!}{%
    \begin{tabular}{ccccccccccccc}
        \toprule
        & & & & & \textbf{LSL} & &
        & & & \textbf{GLSL} & &
        \\
        \addlinespace
        \textbf{Method} & \textbf{Representation} & \textbf{Generation} &
        $K=1$ & 2 & 5 & 10 & 20 & $K=1$ & 2 & 5 & 10 & 20 \\
        \midrule
        Baseline & ResNet-10 \cite{he2016deep} & - & 
        38.5 & 51.2 & 64.7 & 71.6 & 76.3 &
        40.6 & 49.8 & 64.3 & 72.1 & 76.7
        \\
          & SN \cite{koch2015siamese} & - & 
        38.9 & - & 64.6 & - & 76.4 &
        48.7 & - & 68.3 & - & 73.8
        \\
          & MAML \cite{finn2017model} & - & 
        39.2 & - & 64.2 & - & 76.8 &
        49.5 & - & 69.6 & - & 74.2
        \\
          & PN \cite{snell2017prototypical} & - & 
        39.4 & 52.2 & 66.6 & 72.0 & 76.5 &
        49.3 & 61.0 & 69.6 & 72.8 & 74.7
        \\
          & MN \cite{vinyals2016matching} & - & 
        43.6 & 54.0 & 66.0 & 72.5 & 76.9 &
        54.4 & 61.0 & 69.0 & 73.7 & 76.5
        \\
          & PMN \cite{wang2018low} & - & 
        43.3 & 55.7 & 68.4 & 74.0 & 77.0 &
        55.8 & 63.1 & 71.1 & 75.0 & 77.1 
        \\
         & AWG \cite{gidaris2018dynamic} & - & 
        46.0 & 57.5 & \second{69.2} & 74.8 & 78.1 &
        \second{58.2} & \second{65.2} & \second{72.7} & \first{76.5} & \first{78.7}
        \\
        \midrule
        FH \cite{hariharan2016low} & ResNet-10 & LR w/ A. & 
        40.7 & 50.8 & 62.0 & 69.3 & 76.4 &
        52.2 & 59.7 & 68.6 & 73.3 & 76.9
        \\
        P-FH & PN & LR w/ A. & 
        41.5 & 52.2 & 63.5 & 71.8 & 76.4 &
        53.6 & 61.7 & 69.0 & 73.5 & 75.9 
        \\
        FI \cite{wang2018low} & PN & meta-learned LR & 
        45.0 & 55.9 & 67.3 & 73.0 & 76.5 &
        56.9 & 63.2 & 70.6 & 74.5 & 76.5
        \\
          & PMN & meta-learned LR & 
        45.8 & 57.8 & 69.0 & 74.3 & 77.4 &
        57.6 & 64.7 & 71.9 & 75.2 & 77.5
        \\
        \midrule
        \model{} & ResNet-10 & \texttt{cCov-GAN} & 
        \second{47.1} & 57.9 & 68.9 & \second{76.0} & \second{79.3} &
        52.1 & 60.3 & 69.2 & 72.4 & 76.8
        \\
         (Ours) & PN & \texttt{c-GAN} & 
        38.6 & 51.8 & 64.9 & 71.9 & 76.2 &
        49.4 & 61.5 & 69.7 & 73.0 & 75.1
        \\
          & PN & \texttt{cCyc-GAN} & 
        42.5 & 54.6 & 66.7 & 74.3 & 76.8 &
        57.6 & 65.1 & 72.2 & 73.9 & 76.0
        \\
          & PN & \texttt{cDeLi-GAN} & 
        46.0 & \second{58.1} & 68.8 & 74.6 & 77.4 &
        58.0 & 65.1 & 72.4 & 74.8 & 76.9
        \\
          & PN & \texttt{cCov-GAN} & 
        \first{48.4} & \first{59.3} & \first{70.2} & \first{76.5} & \first{79.3} &
        \first{58.5} & \first{65.8} & \first{73.5} & \second{76.0} & \second{78.1}
        \\
        \bottomrule
    \end{tabular}
    }
    \\ \scriptsize
    LR w/ A.: Logistic Regressor with Analogies.
\end{table}

For comparisons, we include numbers reported in previous works under the same
experimental settings. Note that the results for MAML and SN are reported using their published codebases online.
We decompose each method into stage-wise operations for
breaking performance gain down to detailed choices made in each stage.

We provide four models constructed with different GAN choices as justified in Section \ref{sec:isst}. All of our introduced \model{} approaches are trained with NBS-S which would be further investigated with ablation in the next subsection. Results are shown in Table \ref{tab:imagenet}. Our best method
consistently achieves significant improvement over the previous augmentation-based
approaches for different values of $K$ under both LSL and GLSL settings,
achieving almost 2${\%}$ performance gain compared to baselines. We also notice that apart
from overall improvement, our best model achieves its largest boost (\~{}$9\%$) at the lowest
shot over naive baseline and $2.6{\%}$ over Feature Imagination (FI) \cite{wang2018low} under the LSL setting, 
even though we use a simpler embedding technique (PN compared to their PMN). We
believe such performance gain can be attributed to our advanced generation
methods since at low shots, FI applies \textit{discrete}
transformations that its generator has previously learned while we can
now sample through a \textit{smooth} distribution combining all
related base classes' covariance information.

Note that in the LSL setting, all generative methods
assume we still have access to original base examples when learning final
classifiers while non-generative baselines usually don't have this constraint.


\subsection{Discussions} \label{sec:disc}
In this subsection, we carefully examine our design choices for the final version of our \model{}. We
start by unpacking performance gain over the standard batch sampling procedure and
proceed by showing both quantitative and qualitative evaluations on generation
quality.

\begin{figure}[tp]
    \centering
    \begin{subfigure}[b]{0.33\textwidth}
        \includegraphics[width=\textwidth]{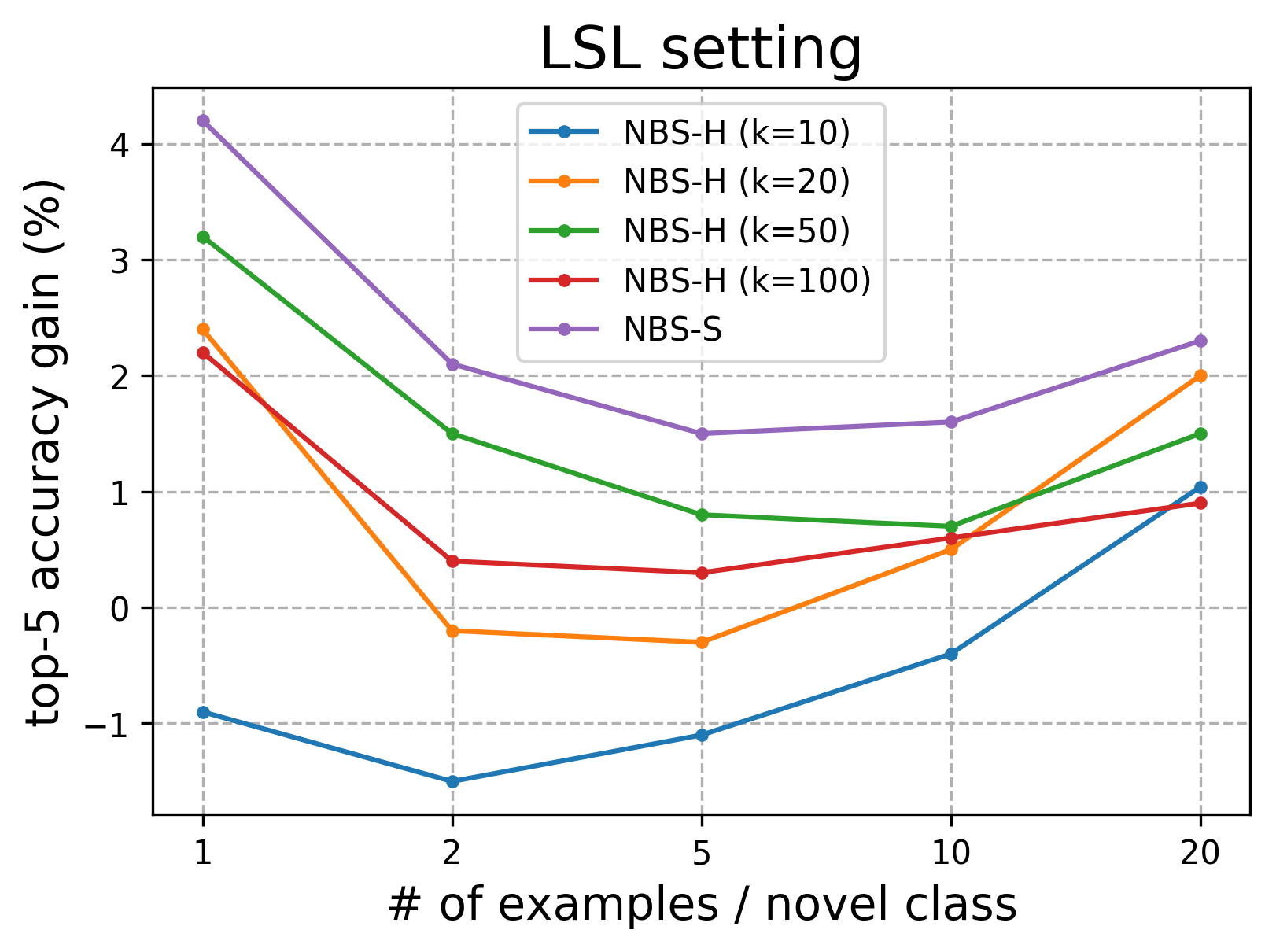}
        \caption{}
        \label{fig:nbs}
    \end{subfigure}
    \begin{subfigure}[b]{0.33\textwidth}
        \includegraphics[width=\textwidth]{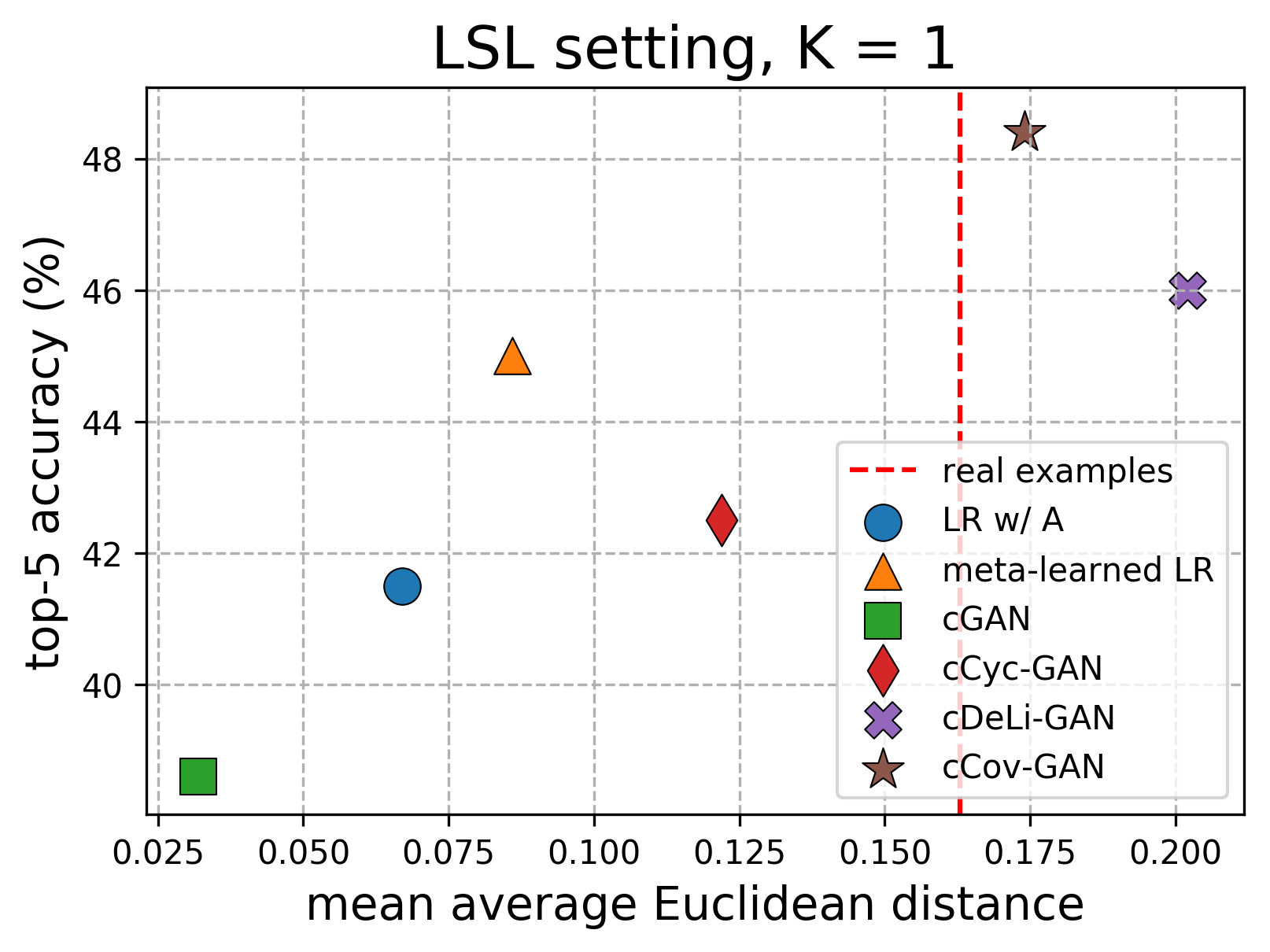}
        \caption{}
        \label{fig:quant}
    \end{subfigure}
    \\
    \begin{subfigure}[b]{0.23\textwidth}
        \includegraphics[width=\textwidth]{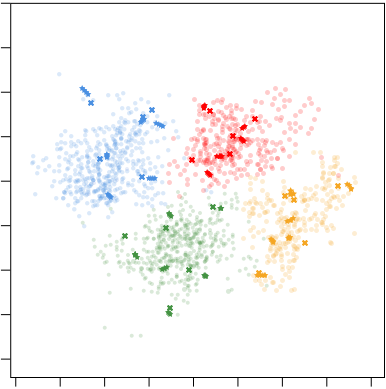}
        \caption{}
        \label{fig:viz-log}
    \end{subfigure}
    \begin{subfigure}[b]{0.23\textwidth}
        \includegraphics[width=\textwidth]{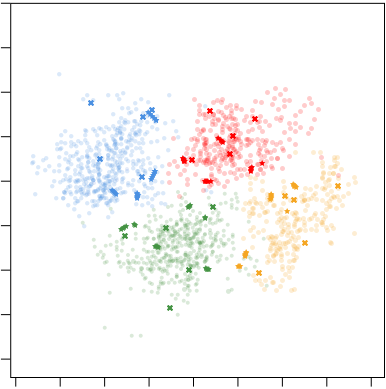}
        \caption{}
        \label{fig:viz-meta}
    \end{subfigure}
    \begin{subfigure}[b]{0.23\textwidth}
        \includegraphics[width=\textwidth]{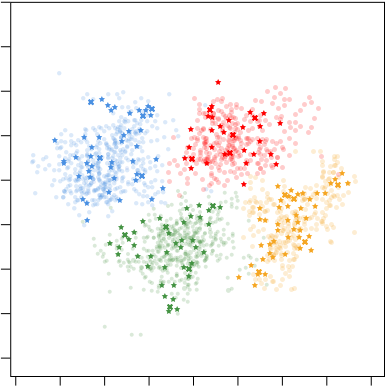}
        \caption{}
        \label{fig:viz-cov}
    \end{subfigure}
    \caption{\textbf{Ablation analysis.}
    \textbf{(a):} Unpacked performance gain for each NBS strategy; \textbf{(b):}
    accuracy vs. diversity; \textbf{(c, d):} Feature Hallucination and Feature
    Imagination lack diverse modes; \textbf{(e):} our best method could
    synthesize both diverse and realistic embeddings. Results are best viewed in
    color with zoom.}
    \label{fig:disc}
\end{figure}

\textbf{Ablation on NBS} \quad
To validate the effectiveness of the NBS strategy over standard batch sampling for
feature augmentation, we conduct an ablation study to show our absolute performance
gain in Figure \ref{fig:nbs}. In
general, we empirically demonstrate that applying NBS improves the performance of low-shot recognition. We also show that the performance of NBS-H is sensitive to the hyper-parameter $k$ in the
$k$-nearest neighbor search. Therefore, the soft assignment is preferable if computational resources allow.

\textbf{Quantitative Generation Quality} \quad
We next quantitatively evaluate the generation quality of the variants introduced
in Section \ref{sec:isst} and previous works as shown in Figure \ref{fig:quant}.
Note that for FH, we used their published codebase online; for FI, we implemented 
the network and train with the procedure described in the original paper.
We measure the diversity of generation via the mean average pairwise Euclidean distance of
generated examples within each novel class. We adopt same augmentation strategies as used for ImageNet
experiments. For reference, the mean average Euclidean distance over real
examples is $0.163$. In summary, the results are consistent with our expectation and support our design choices. Feature
Hallucination and Imagination show less diversity compared to real data. Naive \texttt{c-GAN} even under-performs those baselines
due to the mode collapse. Cycle-consistency and Gaussian mixture noise
do help generation in both accuracy and diversity. However, they either under-
or over-estimate the diversity. Our
covariance-preserving objective leads to the best hallucination quality, since the generated distribution more closely resembles the real data diversity. Another insight from Figure \ref{fig:quant} is that not surprisingly, under-estimating data diversity is more detrimental to classification accuracy than over-estimating.

\textbf{Qualitative Generation Quality} \quad
Figure \ref{fig:viz-log}, \ref{fig:viz-meta}, \ref{fig:viz-cov} show t-SNE
\cite{maaten2008visualizing} visualizations of the data generated by
Feature Hallucination, Feature Imagination and our best model in the prototypical
feature space. We fix the number of examples per novel class $K=5$ in all cases and
plot their real distribution with translucent point clouds. The 5 real examples are
plotted in crosses and synthesized examples are denoted by stars. Evidently, naive
generators could only synthesize novel examples that are largely pulled
together. Although t-SNE might visually drag similar high dimensional points
towards one mode, our model shows more diverse generation results that are better aligned with the latent
distribution, improving overall recognition performance by spreading seed examples in meaningful directions.

\section{Conclusion}

In this paper, we have presented a novel approach to low-shot learning that augments data for novel classes by training a cyclic GAN model, while shaping intra-class variability through similar base classes. We introduced and
compared several GAN variants in a logical process and demonstrated the increasing performance of each model variant. Our
proposed model significantly outperforms the state of the art on the challenging ImageNet benchmark in various settings.
Quantitative and qualitative evaluations show the effectiveness of our method in generating
realistic and diverse data for low-shot learning, given very few examples.

\textbf{Acknowledgments} This work was supported by the U.S. DARPA AIDA Program No. FA8750-18-2-0014. The views and conclusions contained in this document are those of the authors and should not be interpreted as representing the official policies, either expressed or implied, of the U.S. Government. The U.S. Government is authorized to reproduce and distribute reprints for Government purposes notwithstanding any copyright notation here on.


\bibliographystyle{unsrt}
\bibliography{bibliography}


\clearpage
\appendix

\section{Details about Neighborhood Batch Sampling}

In this section, we cover more details in regard to Neighborhood Batch 
Sampling (NBS). We have
considered two instantiations of the translation mapping $R$ and similarity scores
$\alpha$, based on hard $k$-nearest neighbor search and soft selection, respectively. 
Given a novel class $y_n$, we want to select the base classes $\{y_b\}$ that are semantically similar to the $y_n$ query.

\textbf{Hard assignments (NBS-H)}
This sampling method retrieves $k$ uniformly weighted nearest base classes. NBS-H can be formulated as follows,

\begin{equation}
    R(y_n) = 
    \argminn_{\cY_b' \subset \cY_b, |\cY_b'| = k}
    \sum_{y_b \in \cY_b'}
    \| \vl_{y_b} - \vl_{y_n} \|_2^2
    ,\quad
    \alpha(y_b, y_n) = \frac1k\, \forall y_b \in R(y_n).
\end{equation}

Similar heuristics are used in previous works \cite{hariharan2016low,wang2018low} as well
 by introducing a new hyper-parameter $k$. Though NBS-H may save computational
 resources, in
practice, we find it too sensitive to the selection of $k$. In addition to that, 
it treats all
selected base classes as equally related to the target novel class $y_n$, 
which slows the convergence and hurts the performance.

\textbf{Soft assignments (NBS-S)} In this case, all base classes are considered, and weighted by the softmax score over the learned metrics,

\begin{equation}
    R(y_n) = \cY_b
    ,\quad
    \alpha(y_b, y_n) = 
    \frac{\exp{(-\| \vl_{y_b} - \vl_{y_n} \|_2^2)}}
    {\sum_{y_b' \in \cY_b} \exp{(-\| \vl_{y_b'} - \vl_{y_n} \|_2^2})}.
\end{equation}

Through the ablation study, we showed that this batch sampling technique is more effective than 
NBS-H given enough computational resources.

\section{Details about Intermediate GAN Objectives}
In this section, we formulate our full
objectives for intermediate variants derived for the imbalanced set-to-set
translation.

\paragraph{\texttt{c-GAN}}
Its full objective could be defined as a basic minimax game,

\begin{equation}
    G_n^* = \argmin_{G_n} \max_{D_n}
    \cL_{\text{adv}}(G_n, D_n, \cB, \cN).
\end{equation}

\paragraph{\texttt{cCyc-GAN}}
Accordingly, its full objective can be directly derived from cycle-consistency,

\begin{equation}
    \label{eq:ccyc-gan}
    \begin{aligned}
        G_n^* = \argmin_{G_n, G_b} \max_{D_n, D_b}
        \cL_{\text{adv}}(G_n, D_n, \cB, \cN)
        + \cL_{\text{adv}}(G_b, D_b, \cN, \cB) 
        + \lambda_{\text{cyc}} \cL_{\text{cyc}}(G_n, G_b).
    \end{aligned}
\end{equation}

\section{Details about Computing Subgradient of Ky Fan $m$-norm}
\paragraph{Theorem 1} \textit{Given a matrix $\vX$ and its Ky Fan $m$-norm
$\|[\vX]_m\|_* = \sum_{i} \sigma_i (\tilde{\vX})$ where $\tilde{\vX} = \vU
\vSigma \vV^T$ is the $m$-truncated SVD and $\sigma_i
(\cdot)$ is the $i$-th largest singular value, we have,}

\begin{equation}
    \frac{\mathrm{d} \|[\vX]_m\|_*}{\mathrm{d} \vX} = \vU \vV^T
\end{equation}

\paragraph{Proof} Rewrite Ky Fan $m$-norm by its sub-differential set,
\begin{equation}
    \begin{gathered}
        \|[\vX]\|_* = \tr (\vSigma) = \tr (\vSigma \vSigma^{-1} \vSigma)
    \end{gathered}
\end{equation}

Then,
\begin{equation}
    \label{eq:star}
    \mathrm{d} \|[\vX]_m\|_* = \tr (\vSigma \vSigma^{-1} \mathrm{d} \vSigma)
\end{equation}

Since we have,
\begin{equation}
    \mathrm{d} \vX = \mathrm{d} \vU \vSigma \vV^T +
    \vU \mathrm{d} \vSigma \vV^T +
    \vU \vSigma \mathrm{d} \vV^T
\end{equation}

Therefore,
\begin{equation}
    \begin{gathered}
        \vU \mathrm{d} \vSigma \vV^T = \mathrm{d} \vX -
        \mathrm{d} \vU \vSigma \vV^T -
        \vU \vSigma \mathrm{d} \vV^T
        \\
        \Rightarrow 
        \mathrm{d} \vSigma = \vU^T \mathrm{d} \vX \vV -
        \vU^T \mathrm{d} \vU \vSigma -
        \vSigma \mathrm{d} \vV^T \vV
    \end{gathered}
\end{equation}

By the diagonality of $\vSigma$ and anti-symmetricity of $\vU$, $\vV$,
\begin{equation}
    \begin{gathered}
        \vU^T \mathrm{d} \vU \vSigma + \vSigma \mathrm{d} \vV^T \vV = 0
        \\
        \Rightarrow 
        \mathrm{d} \vSigma = \vU^T \mathrm{d} \vX \vV
    \end{gathered}
\end{equation}

Substitute it into Equation \ref{eq:star},
\begin{equation}
    \begin{gathered}
        \mathrm{d} \|[\vX]_m\|_* = \tr (\vSigma \vSigma^{-1} \mathrm{d} \vSigma)
        = \tr (\vU^T \mathrm{d} \vX \vV) = \tr (\vU^T \vV \mathrm{d} \vX)
        \\
        \Rightarrow 
        \frac{\mathrm{d} \|[\vX]_m\|_*}{\mathrm{d} \vX} = \vU \vV^T
    \end{gathered}
\end{equation}
$\qed$

\end{document}